\documentclass[runningheads]{llncs}

\usepackage{eccv}

\usepackage{eccvabbrv}

\usepackage{graphicx}
\usepackage{booktabs}
\usepackage{xcolor}
\usepackage{amsmath}
\usepackage{amssymb}
\usepackage{array}
\usepackage{multirow}
\usepackage{color}
\usepackage{colortbl}
\usepackage{framed}
\usepackage{bm}
\usepackage{bbm}
\usepackage{xspace}
\usepackage{enumitem}
\usepackage{xparse}
\usepackage{url}
\usepackage{wrapfig}
\usepackage{svg}

\usepackage[accsupp]{axessibility}  % Improves PDF readability for those with disabilities.

\usepackage[pagebackref,breaklinks,colorlinks,citecolor=eccvblue]{hyperref}
\usepackage{orcidlink}
\usepackage[dvipsnames]{xcolor}

\def \ie {\emph{i.e.}\xspace}

\def \etal {\emph{et al.}\xspace}

\usepackage{wrapfig} % for wrapping the figure
\usepackage{booktabs} % for tabs spacing
\usepackage{amsmath} % for mathematical features
\usepackage{amsfonts} % for mathematical fonts
\usepackage{courier}
\usepackage{multirow} 
\usepackage{listings} % for code gpt

\usepackage{pgfplots, filecontents}
\pgfplotsset{compat=1.17} % Adjust the version as needed

\newcommand{\tit}[1]{\smallbreak\noindent\textbf{#1.}}

\newcommand{\HEAD}{{HEaD}\xspace}  % for acronym
\newcommand{\PFItc}{$\text{PFI}_{t_c}$\xspace}  % for acronym 
\newcommand{\PFIT}{$\text{PFI}_{\mathcal{T}}$\xspace}  % for acronym 
\newcommand{\CT}{\mathcal{T}}  % for acronym 
\newcommand{\tlast}{t_{c_k}}  % for acronym 
\newcommand{\perfect}{\textit{complete}\xspace}

\usepackage{lipsum} % for dummy text

\usepackage[frozencache=true,cachedir=minted-cache]{minted}
\usepackage[ruled,vlined]{algorithm2e}

\definecolor{commentcolor}{RGB}{110,154,155}   % Define comment color
\definecolor{Gray}{gray}{0.6}
\definecolor{LightCyan}{rgb}{0.88,0.95,1}
\definecolor{blond}{rgb}{0.98, 0.94, 0.75}

\begin{document}
\sloppy

% ---------------------------------------------------------------
\title{Optimizing Resource Consumption\\ in Diffusion Models\\ through Hallucination Early Detection}

\titlerunning{Optimizing Resource Consumption in DM through HEaD}

% Include the authors' ORCID for the camera-ready version, if at all possible.
\author{Federico Betti\inst{1}\orcidlink{0009-0007-3839-0699} \and
Lorenzo Baraldi\inst{2}\orcidlink{0009-0000-4658-8928} \and
Lorenzo Baraldi\inst{3}\orcidlink{0000-0001-5125-4957} \and \\
Rita Cucchiara\inst{3}\orcidlink{0000-0002-2239-283X} \and
Nicu Sebe\inst{1}\orcidlink{0000-0002-6597-7248}}

\authorrunning{F. Betti et al.}

\institute{University of Trento, Italy \email{\{federico.betti,nicu.sebe\}@unitn.it} \and
University of Pisa, Italy \email{lorenzo.baraldi@phd.unipi.it} \and
University of Modena and Reggio Emilia, Italy \email{\{lorenzo.baraldi,rita.cucchiara\}@unimore.it}}

\maketitle

\begin{abstract}
Diffusion models have significantly advanced generative AI, but they encounter difficulties when generating complex combinations of multiple objects. As the final result heavily depends on the initial seed, accurately ensuring the desired output can require multiple iterations of the generation process. This repetition not only leads to a waste of time but also increases energy consumption, echoing the challenges of efficiency and accuracy in complex generative tasks. To tackle this issue, we introduce \HEAD (Hallucination Early Detection), a new paradigm designed to swiftly detect incorrect generations at the beginning of the diffusion process. The \HEAD pipeline combines cross-attention maps with a new indicator, the Predicted Final Image, to forecast the final outcome by leveraging the information available at early stages of the generation process. We demonstrate that using \HEAD saves computational resources and accelerates the generation process to get a \perfect image, \ie an image where all requested objects are accurately depicted. Our findings reveal that \HEAD can save up to 12\% of the generation time on a two objects scenario and underscore the importance of early detection mechanisms in generative models.
\end{abstract}    
\section{Introduction}
\label{sec:intro}
In the rapidly evolving domain of AI, generative models have emerged as a notable subfield, demonstrating an exceptional ability to generate complex visual and textual content~\cite{dhariwal2021diffusion, podell2023sdxl}. The advent of Text-to-Image (T2I) generation marked a significant leap in this domain through the introduction of GAN-based approaches~\cite{goodfellow2014generative, reed2016generative, zhang2017stackgan} and further advancements through large-scale pre-trained Diffusion Models (DM) such as Stable Diffusion (SD)~\cite{rombach2022high} and others~\cite{ramesh2022hierarchical, chefer2023attendandexcite}. These approaches have been instrumental in shaping the generative AI landscape, delivering images that are increasingly indistinguishable from real ones.

\begin{figure}[t]
    \centering
    \includegraphics[width=\textwidth]{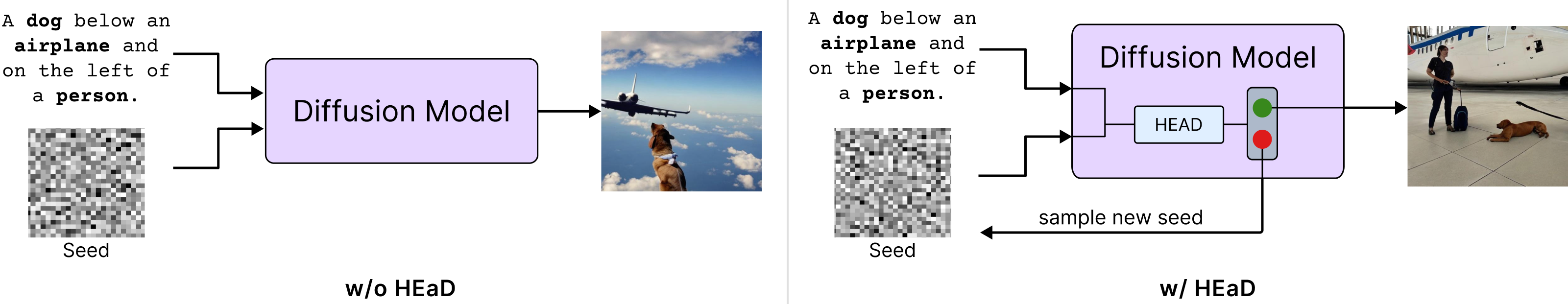}
    \caption{
    Overview of the \HEAD pipeline: during the generation process, \HEAD assesses whether all designated objects will be accurately represented in the final image, determining if the generation process should continue or be restarted with a different seed.
    }
    \label{fig:first_page}
    \vspace{-.3cm}
\end{figure}

Generative models, while progressing, often hallucinate ``long-tail'' objects, which are underrepresented elements in training datasets, and have significant shortcomings when generating multiple objects~\cite{zhang2023deep,samuel2023generating,chefer2023attendandexcite}. Furthermore, they frequently hallucinate attributes, counts, and semantic object relations, which is especially problematic when tasked with rendering scenes involving multiple objects~\cite{liu2023compositional, feng2023trainingfree}.

The challenge is further intensified when generating combinations of specific objects, where diffusion patterns often produce inconsistencies, significantly impacting the quality of the output~\cite{helbling2023objectcomposer}. The choice of the initial seed, which dictates the initial latent noise, is fundamental in navigating the latent space for correct image generation. The dependency on seed selection highlights the unpredictability and variability of these models~\cite{karthik2023dont, samuel2023norm, liu2023compositional, wu2023harnessing, chefer2023attendandexcite}. 
Although automatic evaluation mechanisms could offer a potential solution to these challenges, their adoption is not straightforward. While some attempts in this direction have been made~\cite{betti2023let, lu2023llmscore}, indeed, they still fail at ensuring a sufficiently fast and reliable evaluation.
However, employing these automatic evaluations tends to be slow and resource-intensive. This is largely due to the numerous incorrect results, which require images to be regenerated, thus escalating both time and resource costs.

Addressing these challenges, we introduce \HEAD (Hallucination Early Detection), the first approach designed to enhance both the efficiency and accuracy of generative DMs. \HEAD incorporates the use of cross-attention maps to examine the relationship between the prompt and the internal attention layers of the model, along with the \emph{Predicted Final Image} (PFI) - a prediction of the expected outcome at intermediate stages of the generation process. The combination of PFIs and cross-attention maps allows for the early identification of potential errors by predicting the presence of objects requested by the initial prompt. By preemptively detecting these anomalies, \HEAD hints at stopping the generation diffusion process, thereby conserving resources and reducing the time spent on generating images that would not eventually meet quality standards.
Aiming for a \perfect generation -- where all requested objects are accurately depicted -- halting the generation process based on a prediction of the final outcome proves to be far more efficient than completing an image generation and subsequently evaluating it. This approach not only streamlines the generation process but also enhances resource utilization by avoiding the production and evaluation of substandard images.

We trained two types of networks, each with a different backbone for handling PFI data, followed by CNN-based processing of cross-attention maps. This training occurred at different points in the generation pipeline to assess their impact on prediction quality and potential time savings. Results indicate that using a Visual Transformer as a backbone yields superior outcomes. Moreover, while networks trained towards the later stages of the generation pipeline benefit from higher-quality input and thus demonstrate better performance, those trained earlier exhibit greater potential for time and resource savings. It is also important to note that the methodologies and models described in this study are model-agnostic, \ie they can be seamlessly adapted to any diffusion-based generative model.

In this work, we focus on specific hallucinations: the omission in the generated image of one or more target objects indicated in the textual prompt. We propose both a detector (trained on a dataset of corrected and hallucinated generated images) and a general approach for time-saving prediction that accounts for both the hallucination probability of the specific generative model and the accuracy of the detector. 
For instance, when generating images with prompts involving two objects in non-trivial combinations, SD2 produces hallucinations or missing object errors in 41\% of cases, according to our dataset. Our \HEAD approach can detect the majority of these errors with minimal time overhead, thus saving up to 12\% of the average generation time in this simple scenario.

To sum up, our main contributions are as follows:
\begin{itemize}
    \item We introduce a new element, PFI, and demonstrate that its integration with cross-attention maps effectively facilitates the early detection of objects within generated images.
    \item We propose a comprehensive framework for time saving evaluation. We demonstrate the potential time and resource saving for the generation of \perfect images from multi-object prompts, without compromising the output integrity and generation quality.
    \item A novel classifier for Hallucination Early Detection has been developed that, when integrated into the diffusion process, combines information at each diffusion level and acts as an early evaluator. This classifier stops the process if a hallucination is detected, thereby enhancing the efficiency and accuracy of the generation. 
\end{itemize}

\section{Related Works}
\label{sec:related}

\tit{Text-to-Image Evaluation}
Quantifying the alignment between the generated image and the initial prompt is a challenging task, and as of now, no effective solutions have been identified. Among the assessment metrics,  CLIPScore~\cite{hessel2021clipscore} evaluates the cosine similarity between the prompt and the image, both having undergone processing through their respective visual and textual CLIP backbones. Recent studies~\cite{betti2023let, lu2023llmscore} have proposed innovative scoring mechanisms that leverage the capabilities of Large Language Models (LLMs) and Visual Question Answering. In alignment with this research trajectory, various investigations~\cite{hu2023tifa,Yarom2023whatyou} have introduced diverse methodologies, positioning their work within the reasoning paradigm facilitated by LLMs.

Despite their success in identifying hallucinatory elements in generative models, these works still require the generated image as input, which is produced only in the final step of the diffusion process. 
Additionally, they incorporate evaluation steps beyond image generation, resulting in delays due to the reliance on foundational models within the evaluation pipeline. Conversely, \HEAD enables the detection of hallucinations during the generative process itself, preventing the creation of images that do not align with their prompts.

\tit{Attention Maps in Image Generation}
The integration of attention mechanisms has been a cornerstone in improving image synthesis within generative models. Notably, Chefer \etal refines these processes to enhance image detail~\cite{chefer2023attendandexcite}. Cross-attention layers~\cite{rombach2022high} have significantly boosted visual fidelity, a concept further explored by Hertz \etal to maintain coherence between text prompts and visual outputs~\cite{hertz2022prompttoprompt}. The importance of semantic layouts in improving image quality and interpretability has also been highlighted by Wand \etal ~\cite{wang2022semantic}.
Building on these ideas, SynGen was introduced~\cite{rassin2023linguistic}, which aligns attention maps with prompt syntax to improve attribute correspondence, optimizing the generation process without the need for model retraining. Furthermore, Mao \etal developed a novel method for controlling image synthesis by editing initial noise images, demonstrating that manipulating pixel blocks in initial latent images can influence specific content generation~\cite{Mao_2023}. Additionally, Balaji \etal proposed eDiff-I, an ensemble of expert denoisers that enhances text alignment and visual quality by specializing models for different stages of synthesis~\cite{balaji2022ediffi}.

Following the consensus on the effectiveness of cross-attention as a telltale sign of the fidelity of the generation, our work exploits this information as a discriminating factor for the accurate generation of the final image.

\tit{Seed Importance}
In  Text-to-Image generation, images are significantly impacted by the initial state or starting seed of the diffusion process. 
Indeed, different seeds produce completely different image results as highlighted by Karthik \etal, which claims to generate better-aligned images by evaluating multiple seeds~\cite{karthik2023dont}. 
Furthermore, image editing by directly manipulating the initial noise instead of steering the generation process with additional mechanisms has also been proposed~\cite{Mao_2023,samuel2023norm}. 

Seed selection has gained relevance in the generation of long-tail concepts~\cite{zhang2023deep}. Samuel \etal propose that, in the generation of rare subjects, training predominantly involves exposure to a limited segment of the initial noisy latent space~\cite{samuel2023generating}. This selective exposure during training contributes to the generation of unsatisfactory outcomes across a majority of generative seeds at inference time. Hence, the exploration of diverse generative seeds remains a critical aspect in enhancing generative outcomes. 
To mitigate the occurrence of hallucinations, \HEAD suggests altering the seed in the event of detecting hallucinations in the generative process. 
\newpage
\section{Preliminaries}
\label{sec:preliminaries}

\tit{Latent Diffusion Models}
We focus on the Stable Diffusion (SD) model~\cite{rombach2022high}, which leverages the latent space of an autoencoder rather than the conventional pixel image space. The process begins with an encoder $E$ transforming an image $x$ into a latent code $z = E(x)$. 
A decoder then $D$ aims for accurate reconstruction, insisting $D(E(x)) \approx x$. 
Within this framework, a denoising diffusion probabilistic model (DDPM)~\cite{ho2020denoising, sohl2015deep} operates. This model works on the latent space, creating a denoised version of the input latent $z_t$ at each timestep $t$. Notably, SD enhances this process by integrating a conditioning vector $c(y)$, typically derived from a textual prompt $y$ through a CLIP text encoder~\cite{radford2021learning}.
The objective is to minimize the loss function:
\begin{equation}
\label{eq:loss_function}
L = \mathbb{E}_{z \sim E(x), y, \epsilon \sim \mathcal{N}(0,1), t} \left\lVert \epsilon - \epsilon_{\theta}(z_t, t, c(y)) \right\rVert^2
\end{equation}
where $\epsilon_{\theta}$ is a UNet network~\cite{ronneberger2015u} with attention layers that aims to eliminate the added noise, considering the noisy latent $z_t$, timestep $t$, and conditioning encoding $c(y)$. 

To obtain the final image from the denoised latent representation, the last step involves passing the final latent representation through a Variational Autoencoder (VAE) decoder. This decoder, denoted as $D$, translates the latent space back into the pixel space, thus completing the image generation process. The transition from the final latent state $z_0$ to the generated image $x_0$ can thus be described by
\begin{equation}
\label{eq:vae_decoder}
x_0 = D(z_0),
\end{equation}
where $D$ is the VAE Decoder~\cite{Kingma2014} trained to map the latent representations to high-fidelity images. For further details, we refer the reader to~\cite{rombach2022high}.

\tit{Schedulers in Diffusion Models}
In DMs, schedulers are employed to orchestrate the denoising steps, shaping the generation by modulating noise levels. These algorithms enable the transition from a noisy latent representation to a refined image without adversarial training. In our \HEAD approach, we have tailored the scheduler’s function to extract the PFI at intermediate stages. This modification aims to achieve the most accurate representation of the final image during the generation process.

The transition of latents $z_t$ at time step $t$ to another subsequent state $z_{t'}$ is described as follows. The predicted noise $\epsilon_t$ is firstly obtained from the output of the UNet model, and then the new latents $z_{t'}$ are computed through the update function of the scheduler $\Delta$, as
\begin{equation}
\begin{split}
\label{eq:scheduler}
\epsilon_t &= \epsilon_{\theta}(z_t, t) \\
z_{t'} &= \Delta(z_t, \epsilon_t, t, t').
\end{split}
\end{equation}
Here, $\epsilon_t$ is informed by the current latents and time step, and $\Delta$ is the scheduler update function computing the new latents $z_{t'}$ based on the predicted noise $\epsilon_t$. The behavior of $\Delta$ is determined by the specific scheduler chosen, which dictates the complex dynamics of the denoising process.

\begin{figure*}[t]
    \centering
    \includesvg[width=\textwidth]{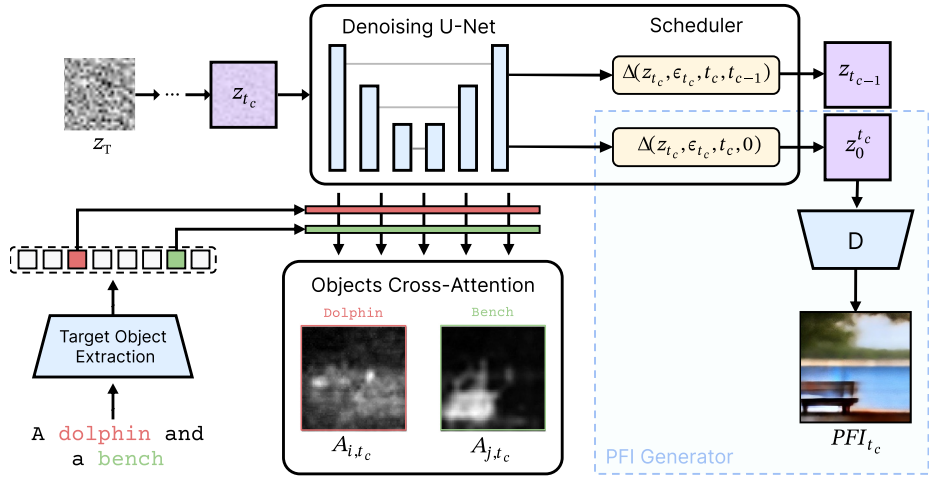}
    \caption{This figure illustrates the process of extracting subjects, cross-attention maps and PFI at each critical timestep \(t_{c} \in \CT\). These elements serve as inputs for the HP network, which evaluates the presence of objects in the final image. For the depicted seed, the bench appears in the final image, whereas the dolphin does not.}
    \label{fig:full_pipeline}
    \vspace{-.3cm}
\end{figure*}
\section{Hallucination Early Detection}
\label{sec:methods}

\HEAD primary goal is to detect and preemptively interrupt faulty generative processes.
Its novelty lies in its ability to perform this detection at intermediate stages of the image generation, leveraging one or more time steps of the diffusion pipeline. Consequently, if the Hallucination Prediction (HP) network predicts that the image will not be \perfect ~-- indicating the absence of at least one target object -- the generative process can be halted and restarted with an alternative seed. This preemptive detection conserves computational resources by preventing the completion of flawed images, eliminating the need to sample a new seed, and avoiding a complete restart from scratch.

In this section, we illustrate the proposed \HEAD approach at inference time to streamline the generation process and, as a result, enable automatic quality assessment of the final output.
The pipeline initial step involves extracting the target objects from the prompt and providing hallucination indicators for the HP network to evaluate.

\subsection{Cross-Attention Maps and PFIs Extraction}
\label{subsec:input_extraction}
Given a prompt $y$ containing a set of target objects $O$ to be generated, the extraction process of these target objects can be formalized as follows:
\begin{equation}
O = \text{TOE}(y)
\end{equation}
where $\text{TOE}(\cdot)$ represents the Target Object Extraction function.
Here, the term ``objects'' refers to words in the prompt directly associated with discernible elements in the image, for which we will extract the corresponding cross-attention maps. While our current methodology primarily focuses on objects, it holds the capability for future expansion to include a wider spectrum of visual concepts, thereby transcending the confines of object-based extraction.

We define a sequence of \textit{critical timesteps}, denoted as $\CT = \{t_{c_1}, \ldots, t_{c_k}\}$, as specific steps in the diffusion process where cross-attention maps and PFI are extracted. These components will serve as inputs for the HP network. 

In diffusion models, the UNet employs cross-attention layers at resolutions from 64 to 8, producing a combined attention map $A_t \in \mathbb{R}^{64 \times 64 \times N}$, where $N$ is the number of tokens from the prompt $y$. For each object $o$ in the target set $O$ and each critical timestep $t_c \in \mathcal{T}$, the specific cross-attention map $A_{o, t_c}$ is derived by filtering $A_t$ for object $o$.

For each critical timestep $t_c \in \CT$, a Predicted Final Image (\PFItc) is extracted. \PFItc represents the prediction of the expected outcome at the end of the generation process, using only information available at timestep $t_c$. In particular, the scheduler projects the latents at $t_c$ to the final step, and the decoder translates these predicted latents into the image space. The process is defined as follows:
\begin{equation}
\begin{split}
\label{eq:pfi_projection}
\epsilon_{t_c} &= \epsilon_\theta(z_{t_c}, t_c) \\
z_{0}^{t_c} &= \Delta(z_{t_c}, \epsilon_{t_c}, t_c, 0) \\
\text{PFI}_{t_c} &= D(z_{0}^{t_c})
\end{split}
\end{equation}
where $\epsilon_{t_c}$ represents the predictive noise obtained from the UNet model at critical timestep $t_c$. The function $\Delta$ updates the latents ${z}_{t_c}$ to the predicted latents at the final timestep, denoted as ${z}_{0}^{t_c}$. Finally, the VAE decoder $D$ translates these predicted final latents into \PFItc. 

\begin{figure*}[t]
    \centering
    \includesvg[width=\textwidth]{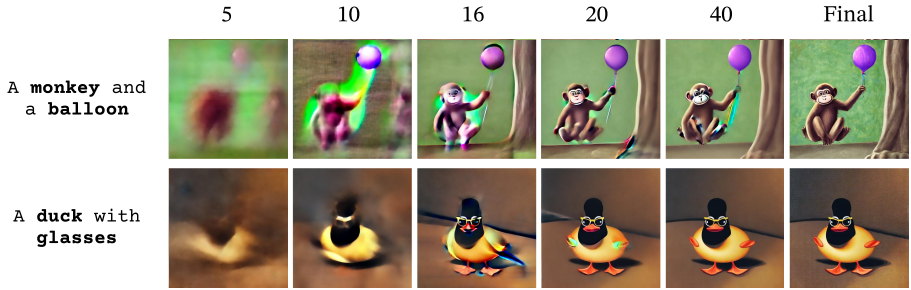}
    \caption{Qualitative examples of the Predicted Final Image for each prompt at different critical timesteps. Already from the 16\textsuperscript{th} step the final image is fully represented and the presence of objects can be predicted.}
    \label{fig:qualitativo_PFI}
    \vspace{-.3cm}
\end{figure*}

Examples of PFIs extracted at different timesteps are shown in Fig.~\ref{fig:qualitativo_PFI}. The collection of PFIs, namely \PFIT, and attention maps, $A_{O, \CT}$, across all critical timesteps provides a comprehensive dataset for the HP network to analyze and predict the presence of objects in the final image.

\subsection{Hallucination Prediction Network}
\label{subsec:hallucination_prediction_network}

During the evaluation phase, the Hallucination Prediction network takes as input the cross-attention maps $A_{o}$ for a specific object and the \PFIT and outputs a binary prediction indicating the presence or absence of that specific target object in the final image:
\begin{equation}
\label{eq:hallucination_prediction}
H_{o} = \text{HP}(A_{o, \CT}, \text{PFI}_{\CT})
\end{equation}
where $H_{o}$ is the binary prediction for object $o$. The training methodology for the HP network is detailed in Section~\ref{sec:training}.

The reliability of the HP network is critical to prevent unnecessary terminations of the generation and ensure that objects that would have been present are not prematurely discarded.

In all network configurations, as feature extractor from cross-attention maps, we utilize a series of four convolutional layers that bring the input from a $1\times64\times64$ to a final $128\times7\times7$ shape.

\tit{HP-R -- HP-V}
For scenarios where the number of critical timesteps selected is 1, \ie $|\CT| = 1$, we have explored two distinct backbone options: Resnet50 and Vision Transformer (ViT), which correspond to the HP-R and HP-V architectures respectively. These architectures are tailored to process the PFI to extract features that are then combined with those extracted from the cross-attention maps to get the final prediction. HP-R concatenates the output of the Resnet backbone on the channel level with the processed cross-attention maps and a further two-layer CNN is used to merge channels down to $512$. HP-V processing outputs a vector that is concatenated to the flattened version of the cross-attention maps output. Both networks employ a final linear layer for the binary prediction.

\tit{HP-MultiR}
For scenarios where $|\CT| > 1$, we have created the HP-MultiR network in which Resnet50 was used to extract features from different timesteps in parallel. Features and the processed cross-attention maps are concatenated on the sequence dimension before applying two Conv3d layers with a final 3D pooling to reduce dimensions. A final classifier is eventually used for the final prediction.

\tit{HP-A}
Additionally, we developed the HP-A network to specifically investigate the influence of attention maps on hallucination prediction. 
This network configuration excludes the PFI from its input, focusing solely on the features extracted from attention maps. A final prediction layer is attached directly to the common cross-attention maps feature extractor.
By employing a similar architecture with convolutional layers as the other HP variants, the HP-A network focuses on evaluating how well attention maps alone can predict hallucinations. The results from this model provide critical insights into how effectively attention maps alone can inform the hallucination prediction process in diffusion models.

\section{Hallucination Network Training}
\label{sec:training}

\tit{Dataset Creation}
\label{subsec:dataset_selection}
To train the Hallucination Prediction network we collected 900 prompts obtained by combining 75 distinct animal subjects with 12 objects with the prompt ``\texttt{A \{animal\} and a \{object\}}''. To augment the dimensionality of the dataset and thoroughly investigate output variations influenced by different seeds, we generated 12 images for each prompt using distinct seeds. Following this protocol we generated nearly 10.000 images by making use of Stable Diffusion v2.0 generator~\cite{rombach2022high}. 
During generation, we fixed 50 steps of the diffusion process and we collected the PFI and cross-attention maps $A$ at multiple time steps\footnote{In particular, the critical steps $\CT$ are chosen as follows: [0, 1, 2, 3, 4, 5, 6, 7, 8, 9, 10, 12, 14, 16, 18, 20, 25, 40]}.

\tit{Target Objects Extraction}
\label{subsec:training_object_extraction}
While our dataset comprises prompts with objects in predetermined positions for simplicity, we integrated object extraction to simulate real-world scenarios. For this purpose, we employed gpt-3.5-turbo-1106~\cite{openai_gpt-4_2023}, selected for its robust zero-shot generalization abilities. This method stands in contrast to conventional text tagging techniques that generally necessitate specific training for each domain.

The extraction procedure is time-efficient and can be executed concurrently with the initial diffusion steps. 
Details on the specific prompts used in this study can be found in the supplementary materials.

\tit{Label Creation}
\label{subsec:label_creation}
An essential feature is the development of an automatic labeling system to confirm the presence of particular objects in the generated images. This system must function without a fixed set of object labels, requiring the adoption of an open vocabulary approach.
To achieve this, we adopted OWLv2~\cite{minderer2023scaling}, an open vocabulary detector renowned for its robust detection capabilities and for providing confidence scores for each identified object. 
\section{Time Saving Analysis}
\label{sec:time_saving}

Our study primarily explores the time-saving benefits of the \HEAD approach in DMs when trying to generate a \perfect image. In our analysis, we found that accurate generation of both objects in complex scenarios was achieved in only 59\% of cases by SD2 without \HEAD. This statistic underscores the challenges models encounter when generating multiple objects accurately, particularly as the complexity of the prompt and object combinations increase. Certainly, with more objects and increasingly complex prompts, the probability of correct generation diminishes, which in turn heightens the impact of HEaD on time-saving.

In the dataset, which is made with 12 seeds per prompt, we found that every prompt successfully led to at least one correctly generated image, and 98.4\% resulted in at least three accurate images. This confirms the feasibility of the prompts for some random seeds. \HEAD serves here as an implicit evaluator, swiftly identifying instances where the generated image is likely to be inaccurate. By promptly halting these less promising generative paths, \HEAD allows for more efficient use of resources, enabling quicker initiation of new generation attempts with different seeds.

\subsection{\HEAD impact on Time Saving}
\tit{HP Performance}
Labels in our dataset are created using an open vocabulary detector, which assesses whether each object is present (1) or absent (0) in the images. The HP network, based on these labels, decides whether to continue or halt the image generation process. When a True Positive (TP) occurs, the correct generation proceeds uninterrupted, having no effect on computation time. Conversely, a False Positive (FP) allows an incorrect generation to continue without interruption, thus missing an opportunity for time savings, but still not impacting computation time. A True Negative (TN) indicates an incorrect generation has been correctly halted, leading to time savings. Finally, a False Negative (FN) means a correct generation is mistakenly stopped, resulting in a loss of time.

Thus, in order to save computational time, the network should be trained to balance both high recall and a high TN-rate. High recall ensures the HP network effectively identifies all instances of correct generation, minimizing FNs and avoiding unnecessary termination of accurate processes. Simultaneously, a high TN-rate boosts the HP network’s capability to maximize true negative outcomes, allowing for early termination of incorrect generations by accurately identifying cases where not all requested objects in the prompt are included.
This dual focus on both recall and TN-rate optimizes the generation process by reducing time loss, yet still maintaining the quality of the output.

\tit{Critical Timesteps selection}
The ratio $\frac{\tlast}{T}$, where $\tlast$ denotes the latest $t$ in the set of critical timesteps $\CT$ and $T$ is the total number of steps in the generation process (with 50 being the standard for SD), plays a crucial role in determining the percentage of time saved. An earlier detection, indicated by a smaller $\tlast$, can potentially lead to greater time savings in case of a correct hallucination identification. However, this scenario presents a significant challenge: in the initial stages the quality of attention maps and PFIs is lower. This lower quality affects the performance of the HP network resulting in reduced recall and TN-rate, as shown in the plot in Fig.~\ref{fig:resnet_comparison}. Therefore, this tradeoff between early detection and maintaining the quality of attention maps and PFIs is essential for maximizing the efficiency of the \HEAD approach.

\begin{figure}[t]
\centering
\begin{minipage}[t]{.49\textwidth}
  \centering
  \includesvg[width=\linewidth]{Images/recall_tn_rate_toy.svg}
  \vspace{-.30cm}
  \caption{Recall and TN-rate values for HP-R across various $\tlast$. Lower $\tlast$ values, associated with lower quality input, significantly impact the TN-Rate but minimally affect Recall. Consequently, the overall time saved tends to be greater for smaller $\tlast$ values.
  }
  \label{fig:resnet_comparison}
\end{minipage}\hfill
\begin{minipage}[t]{.49\textwidth}
  \centering
  \includesvg[width=\linewidth]{Images/time_saved_toy.svg}
  \vspace{-.30cm}
  \caption{Relative time saving between adopting or not the \HEAD approach to reach a \perfect generation, using HP-R with different $\tlast$, depending on the probability of a correct image generation. The vertical red line marks the probability of correct generation in a two-objects scenario, \ie 59\%.}
  \label{fig:resnet_probability_related}
\end{minipage}
\end{figure}

Finally, to quantify the time saved or lost using the model, we conducted Monte Carlo simulations based on the models presented in the next section. The algorithm calculates a savings of $\frac{\tlast}{T}$ of the generation time when a true negative (TN) is detected. Conversely, it accounts for a time loss when a new restart is necessary due to a false negative (FN).
The detailed algorithm and simulation results are provided in the supplementary materials.

\section{Experimental Results}
\label{sec:results}
The evaluation of various HP Network variants underscores their influence on the image generation process. The computed Recall and TN-rate metrics, which are influenced by the $\tlast$ value, serve as key indicators of model performance. As depicted in Fig. \ref{fig:resnet_comparison}, the TN Rate typically increases with a higher $\tlast$, whereas Recall tends to remain stable across different stages of detection.

To provide a final efficiency assessment, the percentage of time saved during generation has been adopted as the primary metric for final comparison. This metric integrates the effects of varying $\tlast$, Recall, and TN-Rate values, offering a quantifiable measure of each model’s effectiveness in reducing generation times. For these experiments, a correct generation probability of 59\%, as derived from the dataset, has been employed to ensure accuracy in the evaluations.
Table \ref{tab:results_ts_comparison} provides a comparative analysis of HP-R and HP-V, illustrating the time saved when these networks operate at different $\tlast$ intervals. Both networks have the highest impact when $\tlast = 8$, where HP-V saves up to 12.66\% of generation time.
Higher $\tlast$ values can enhance input quality and metric results, but they may limit time-saving opportunities. No models bring any benefit when using $\tlast \ge 25$, as the time saved in case of a correct prediction is insufficient. 

In Fig. \ref{fig:resnet_probability_related}, an analysis is presented to illustrate the relationship between the relative time saved and the generation probability across different $\tlast$ values. The vertical line indicates a 59\% correct generation probability, typical for scenarios involving two objects, as observed in our dataset. More complex prompts, which often require synthesizing additional objects, tend to have lower probabilities of achieving a \perfect generation, thus enhancing potential time savings. Notably, $\tlast = 8$ offers the optimal balance, providing significant time savings, especially when the probability of \perfect generation is as low as 40\%, where time savings can reach up to 30\%.
Conversely, when the probability of a \perfect generation is high, using $\tlast = 5$ results in considerable time loss due to imperfect Recall, which can prematurely halt a correct generation. Additionally, employing \HEAD at $\tlast = 40$ provides no benefits in any scenario, as the time saved in the rare event of a true negative is merely 20\%, considering the 50 steps generation pipeline of SD2.

In Table~\ref{tab:ablation_model}, HP-A testing serves as an ablation study to underscore the significance of Predicted Final Images. In the absence of PFIs, which are unique per image and not per object, the HP-A model shows a marked decrease in its ability to detect early hallucinations and thus in time saved. With $\tlast = 10$ only 6.65\% of generation time is saved. 

\begin{table}[t]
\centering
\begin{minipage}[t]{.49\textwidth}
  \centering
    \renewcommand{\arraystretch}{1.2} % Adjust the factor for desired interline spacing
  \setlength{\tabcolsep}{2.0em}
  \resizebox{\linewidth}{!}{
  \begin{tabular}{ccc}
    \toprule
    $\CT$ &   \textbf{HP-V} & \textbf{HP-R} \\
    \midrule
     5 &    9.7 &   9.11 \\
     \rowcolor{LightCyan}
     8 &  \textbf{12.66} &  \textbf{10.56} \\
    10 &   9.68 &  10.34 \\
    16 &   6.72 &   8.93 \\
    18 &   5.77 &   5.78 \\
    20 &   5.75 &   7.25 \\
    25 &  -0.35 &   5.32 \\
    40 & -14.11 & -11.67 \\
        \bottomrule
  \end{tabular} 
  }
  \vspace{.3cm}
  \caption{Percentage of time saved for all models. $\tlast$ is the last diffusion timestamp considered over the 50 of SD2.}
  \label{tab:results_ts_comparison}
\end{minipage}\hfill
\begin{minipage}[t]{.49\textwidth}
  \centering
  \renewcommand{\arraystretch}{1.4} % Adjust the factor for desired interline spacing
  \setlength{\tabcolsep}{.5em}
  \resizebox{\linewidth}{!}{
  \begin{tabular}{ccc}
    \toprule
    \textbf{Model} & \textbf{$\CT$} & \textbf{\% Time Saved} \\
    \midrule
    \multirow{3}{*}{HP-A} & 10 & 6.65\% \\
     & 16 & 3.04\% \\
     & 20 & -0.73\% \\
    \midrule
    \multirow{3}{*}{HP-Multi} & 6-8-10 & -3.72\% \\
     & 10-12-14 & 8.99\% \\
     & 16-18-20 & 6.88\% \\
            \bottomrule
  \end{tabular} 
  }
  \vspace{.3cm}
  \caption{Percentage of time saved for HP-A and HP-Multi in different $\CT$ scenarios.}
  \label{tab:ablation_model}
\end{minipage}
\end{table}

The HP-Multi model takes an advanced approach by focusing on multiple $\CT$. A noteworthy aspect of HP-MultiR performance is its effectiveness in later timesteps ($\tlast = 14$), compared to a less marked performance in early timesteps. This discrepancy can be attributed to the inhomogeneity of the data in the early stages, where the characteristics of the data change considerably from one step to the next. This variability makes the mixing of the features in these early stages less effective. In contrast, data in later stages tend to be more uniform and stable, allowing for more effective learning and integration of features from multiple time steps, thus improving model performance.

\section{Conclusions}
\label{sec:conclusions}

This paper introduces \HEAD, an innovative approach that not only enhances the efficiency and accuracy of image generation with Diffusion Models but also significantly reduces computational resources.
A key innovation is the Predicted Final Image, an effective early error prediction indicator when used in conjunction with cross-attention maps.
The effectiveness of our framework in saving time is closely tied to the recall and TN-rate of the Hallucination Prediction network, highlighting \HEAD's capacity to improve image generation in a variety of complex scenarios.

\HEAD represents a preliminary step in exploring the sustainability and effectiveness of diffusion models, especially for large, complex datasets.
Looking ahead, we are committed to further advancing this field of study also by collecting larger datasets with more target objects and more complex visual prompts and proposing challenges for the scientific community to test better early detectors.

\section{Acknowledgment}
This work was supported by the MUR PNRR project FAIR - Future AI Research (PE00000013) funded by the NextGenerationEU, the PRIN project CREATIVE (Prot. 2020ZSL9F9), the EU Horizon projects ``European Lighthouse on Safe and Secure AI (ELSA)'' (HORIZON-CL4-2021-HUMAN-01-03), co-funded by the European Union (GA 101070617) and “ELIAS - European Lighthouse of AI for Sustainability” (No. 101120237).  Further, we thank G. Fiameni (NVIDIA) for helping with the generation of the dataset.

\bibliographystyle{splncs04}
\bibliography{main}
\end{document}

\maketitle

\setcounter{equation}{8}
\setcounter{table}{6}
\setcounter{figure}{5}

%\begin{enumerate}
%\item esempio dal nuovo dataset. Prompt con 5 oggetti dove mostriamo anche l'estrazione dei soggetti (che nel paper non viene utilizzata). (X 2 o 3 prompt, alcuni con errori)
%estrazione dei soggetti
%C.A. Map al timestep 8
%PFI al time step 8
%predicted presence with HEaD

%\item prompt GPT per estrazione oggetti
%\item algoritmo per la simulazione montecarlo. Proviamo a vedere se riusciamo con pseduocode

%\end{enumerate}
In this investigation, we delve into additional details associated with the qualitative examination of the HEaD input, considering additional subjects and diverse prompts. Furthermore, we provide further insights into the Monte Carlo simulations and the processes involved in object extraction.

\section{Additional HEaD input examples}
\begin{figure*}[t]
    \centering
    \includesvg[width=\textwidth]{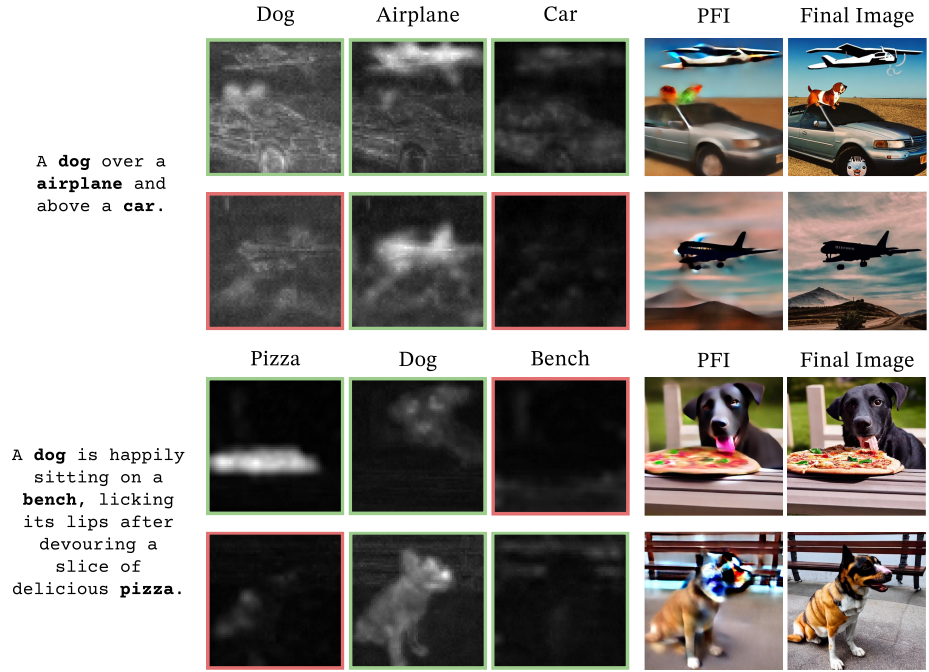}
    \vspace{-.65cm}
    \caption{Examples of Target Objects Extraction, their cross attention map, and the Predicted Final Image at timestep 16. Highlighted with the green border are the cross attention maps with the object in the image, in red otherwise.}
    \label{fig:suppqualitatives}
    \vspace{-.3cm}
\end{figure*}

In our experimental setup, HEaD was employed on prompts featuring two subjects, involving the combination of 75 unique animal subjects with 12 objects. 
Starting from less structured prompts collected by Bakr \etal~\cite{Bakr_2023_ICCV}, we visually analyze our input pipeline in Figure~\ref{fig:suppqualitatives}. 
In these examples, we performed the object extraction pipeline following the procedure detailed in Section~\ref{sec:label_extraction}, and generated the images using Stable Diffusion 1.4~\cite{rombach2022high}.
Notably, first insights on subject hallucinations are still detectable at timestep 16 of the generation process. For instance, considering the prompt \texttt{A dog over a airplane
and above a car}, the second row doesn't represent either the \texttt{dog} or the \texttt{car} in its PFI.  Moreover, the cross-attention maps of these missing subjects are less emphasized compared to the upper row, where all the objects are well represented. 
Similar outcomes are observed in the prompt \texttt{A dog is happily sitting on a bench, licking its lips after devouring a slice of delicious pizza}. Indeed, \texttt{pizza} is missing from both the Final Image and the PFI in the example in the 4$^{\text{th}}$ row. Compared to the 3$^\text{rd}$ instance, where all the subjects are well-represented in the PFI, the cross-attention map is more activated in the case of \texttt{pizza} subject.

%\begin{figure*}[t]
%    \centering
%    \includegraphics[width=\textwidth]{Images/SupplementaryImages/DetectionError.pdf}
%    \vspace{-.65cm}
 %   \caption{Example of wrong Target Object extraction: \texttt{group} and \texttt{lights} are not valid objects to evaluate.}
 %   \label{fig:supp_qualitatives_error}
 %   \vspace{-.3cm}
%\end{figure*}

\section{Monte Carlo HEaD simulations}

The Python pseudocode detailed in Listing~\ref{algo:monte_carlo_pseduocode} simulates the time savings achieved by implementing the \HEAD approach within the image generation process. Its effectiveness depends on the model's performance, particularly in terms of Recall and TN-Rate, and the number of requested subjects $|O|$. \HEAD analyzes each subject independently, and it only requires one of the objects to be predicted as absent to halt the generation and restart with a new seed. The time saving occurs when the model incorrectly generates an image, i.e., a subject is not present, and \HEAD is able to predict this and immediately restart the generation with a different seed. The time saved in each of these instances is dependent on $\tlast$, which represents the maximum critical timestep used for analyzing the cross-attention maps and the PFIs.

\begin{listing*}
\begin{minted}
[
frame=lines,
framesep=2mm,
baselinestretch=1.0,
fontsize=\footnotesize,
linenos
]{Python}
# cgp (complete_generation_probability): the probability of 
#   having an image with all requested objects
# recall: recall of the HP network
# tn_rate: tn_rate of the HP network
# time_per_model_iteration: time for completing a generation
# max_step_used: last step used for HEaD evaluation
# num_objects: number of objects to evaluate
# total_steps: number of generation step, 50 for SD2
# num_simulations: number of Monte Carlo simulations

# Computing time when HEaD model detects failure
time_used_per_TN = (max_step_used / total_steps) * \
    time_per_model_iteration
# Time with HEaD approach
time_with_head = 0
for _ in range(num_simulations):
    success = False
    while not success:
        # Generate an image
        is_image_complete = random.random() < cgp
        if is_image_complete:
            # HP network must predict all success 
            #to stop the generation process
            hp_predicts_success = all(
                random.random() < recall for _ in range(num_objects)
            )
            if hp_predicts_success: # TP
                time_with_head += time_per_model_iteration
                success = True
            else: # FN
                time_with_head += time_per_model_iteration
        else:
            # The generation has at least one object hallucinated. 
            # If HP finds one hallucinated object, 
            # generation is restarted sooner
            hp_predicts_failure = any(
                random.random() < tn_rate for _ in range(num_objects)
            )
            if hp_predicts_failure: # TN
                time_with_head += time_used_per_TN
            else: # FP
                time_with_head += time_per_model_iteration
# Time with HEaD approach
avg_time_with_HEaD = time_with_head / num_simulations
# Time without HEaD approach
avg_time_no_HEaD = time_per_model_iteration / cgp
return 1 - avg_time_with_HEaD / avg_time_no_HEaD
\end{minted}
\caption{Python pseudo code for \HEAD Monte Carlo simulation.}
\label{algo:monte_carlo_pseduocode}
\end{listing*}

\section{Target Objects Extraction}
\label{sec:label_extraction}
As detailed in Section 5, our object extraction process is a critical component of the \HEAD approach. We employed GPT-3.5-turbo-1106~\cite{openai_gpt-4_2023} to recognize and extract entities from text prompts. The entities, in this context, are elements with a physical representation. 

The system was instructed to use a specific prompt to guide its entity recognition process. The prompt used was as follows:

\begin{minipage}{\linewidth}
\small
\texttt{\\You are a system that is able to recognize entities in a text. Entities are objects, people, animals, etc. that have a physical representation. Avoid to include abstract subjects. Do not consider adjectives in the entities.\\}
\end{minipage}
To enhance the model accuracy, we also provided a few-shot learning approach with relevant examples. This method was crucial in ensuring the model's focus on extracting only concrete entities while excluding abstract concepts and adjectives, aligning with the objectives of our research and the operational requirements of the \HEAD pipeline.
Figure~\ref{fig:suppqualitatives} presents examples of Target Object Extraction, wherein the output of the process in both prompts faithfully corresponds to the anticipated subjects.

{
    \small
    \bibliographystyle{splncs04}
    \bibliography{main}
    % \bibliography{bibliography}
}